\documentclass[lettersize,journal]{IEEEtran}
\usepackage{amsmath,amssymb,amsfonts}
\usepackage{array}
\usepackage[caption=false,font=normalsize,labelfont=sf,textfont=sf]{subfig}
\usepackage{textcomp}
\usepackage{stfloats}
\usepackage{url}
\usepackage{verbatim}
\usepackage{graphicx}
\usepackage{cite}
\usepackage[table]{xcolor}
\usepackage{booktabs}
\usepackage{multirow}
\usepackage{makecell}
\usepackage{tabularx}
\usepackage{pifont}
\usepackage[capitalize,nameinlink]{cleveref}
\newcommand{\tabfont}{\footnotesize}
\definecolor{lightgrayrow}{gray}{0.92}
\definecolor{lightbluerow}{RGB}{224,238,255}
\newcommand{\cmark}{\ding{52}}

\begin{document}

\title{TSHA: A Benchmark for Vision-Language Models in Trustworthy Safety Hazard Assessment Scenarios}

\author{Qiucheng Yu, Ruijie Xu, Mingang Chen, Jianfeng Dong, and Xin Tan%
\thanks{Qiucheng Yu is with City University of Hong Kong. Ruijie Xu is with East China Normal University. Mingang Chen is with Shanghai Development Center of Computer Software Technology. Jianfeng Dong is with Zhejiang Gongshang University. Xin Tan is with East China Normal University. Corresponding author information will be completed in the final manuscript.}%
}

\maketitle

\begin{abstract}
Recent advances in vision-language models (VLMs) have accelerated their application to indoor safety hazards assessment. However, existing benchmarks suffer from three fundamental limitations: (1) heavy reliance on synthetic datasets constructed via simulation software, creating a significant domain gap with real-world environments; (2) oversimplified safety tasks with artificial constraints on hazard and scene types, thereby limiting model generalization; and (3) absence of rigorous evaluation protocols to thoroughly assess model capabilities in complex home safety scenarios. To address these challenges, we introduce TSHA (\textbf{T}rustworthy \textbf{S}afety \textbf{H}azards \textbf{A}ssessment), a comprehensive benchmark comprising 66,668 validated question-answer pairs, including 64,961 carefully curated training QA pairs drawn from existing indoor datasets, internet frames/images, AIGC images, newly captured images, and Hunyuan panoramic images. This benchmark also includes a highly challenging test set with 1,707 QA pairs, comprising not only a carefully selected subset from the training distribution but also newly added Sora-generated videos and Hunyuan panoramic images containing multiple safety hazards, used to evaluate the model's robustness in complex safety scenarios. Extensive experiments on 22 popular VLMs demonstrate that current VLMs lack robust capabilities for safety hazard assessment. Importantly, models trained on the TSHA training set achieve a significant performance improvement of up to +18.3 points on the TSHA test set and also exhibit enhanced generalizability across other benchmarks, underscoring the substantial contribution and importance of the TSHA benchmark.
\end{abstract}


\section{Introduction}
\label{sec:intro}
\IEEEPARstart{R}{ecent} progress in Vision-Language Models (VLMs)~\cite{dai2026thinkmatter,chakraborty2026toward,liu2026mtmcs} has revealed their remarkable abilities across diverse practical applications, including visual exploration~\cite{xu2026multi,gou2026mixture}, autonomous driving~\cite{zhang2026vision,gao2026foundation,wang2025physically}, real-world instruction following~\cite{ji2025robobrain,he2026continual}, and embodied navigation~\cite{zhang2025mem2ego,yokoyama2025film,qiu2025salient}. The versatility of VLMs has catalyzed significant research interest, particularly in the emerging domain of automated indoor safety inspection~\cite{gao2025homesafebench,mullen2024don,hassan2024coherence,huang2025framework,yin2024safeagentbench}, where they offer the potential to augment or replace traditional human assessments. However, current VLM-based approaches exhibit critical limitations in home safety hazard assessment, primarily stemming from inadequate domain knowledge about potential domestic risks. Addressing this knowledge gap requires a robust evaluation framework that serves as both a challenging testbed and a rich source of domain knowledge.

Existing evaluation benchmarks for safety inspection fall into two categories: text-based approaches~\cite{yin2024safeagentbench,mullen2024don,lou2026helpers} that convert visual data into textual representations, such as object relationship graphs, for processing by pure LLMs, and vision-based methods~\cite{gao2025homesafebench,hassan2024coherence,huang2025framework} that directly evaluate visual inputs or construct virtual environments for VLM training. These paradigms suffer from three fundamental limitations: (1) over reliance on synthetic data from simulators, causing a substantial reality gap that impedes practical deployment; (2) constrained scenario diversity with artificially limited hazard categories, failing to capture the broad spectrum of real-world safety concerns; and (3) simplistic evaluation protocols using mostly static images or narrow hazard settings, providing inadequate assessment of model generalization.

To address these limitations, we introduce TSHA (Trustworthy Safety Hazards Assessment), a comprehensive benchmark for trustworthy safety hazard assessment. TSHA offers three key advantages. (1) \textbf{Diverse and High-Quality Data}. TSHA aggregates a large-scale dataset containing 66,668 validated QA pairs, including 64,961 training QA pairs, curated from complementary sources: existing indoor datasets, internet frames/images, AIGC images, newly captured images, and Hunyuan panoramas. It integrates realistic videos, synthetic images, and authentic photographs, effectively mitigating the reality gap between prior simulators and the real world. (2) \textbf{Comprehensive Hazard Coverage}. Unlike previous datasets~\cite{gao2025homesafebench,anonymous2025robotrust,huang2025framework} that restrict hazard type diversity, TSHA encompasses 12 parent hazard groups and 64 fine-grained hazard types. Beyond common hazards such as fires and falls, it includes more complex, hard-to-assess risks, for instance, unattended children in proximity to dangerous objects or clutter obstructing emergency escape routes. (3) \textbf{Challenging Test Set}. The benchmark features a meticulously designed test set with 1,707 QA pairs, which not only includes a subset from the training distribution but also supplements it with more challenging data: videos generated by Sora~\cite{brooks2024video} and panoramic images from Tencent Hunyuan world model~\cite{huang2025voyager}. Each video or image contains multiple concurrent safety hazards, requiring models to identify as many hazards as possible simultaneously. A detailed comparison of TSHA against prior work is presented in \Cref{tab:safety_datasets}.

\begin{table*}[t]
\centering
\caption{Comparison between TSHA and existing safety-hazard assessment benchmarks. TSHA reports both parent and fine-grained hazard labels, and uses 58 audited primary scene labels for the source-media inventory.}
\label{tab:safety_datasets}
\tabfont
\setlength{\tabcolsep}{8pt}
\renewcommand{\arraystretch}{1.05}
{%
\begin{tabular}{lccccc}
\toprule
\textbf{Dataset} & \textbf{Samples / QA pairs} & \textbf{Hazard categories} & \textbf{Scene categories} & \textbf{Modality} & \textbf{Source} \\
\midrule
SafetyDetect~\cite{mullen2024don} & 1,000 & 3 & 7 & Text & Simulator \\
M-CoDAL~\cite{hassan2024coherence} & 908 & 16 & -- & Images & Internet \\
SafeAgentBench~\cite{yin2024safeagentbench} & 750 & 10 & 1 & Images & Simulator \\
Safe-BeAl~\cite{huang2025framework} & 2,027 & 8 & 1 & Text & Simulator \\
ROBOTRUST~\cite{anonymous2025robotrust} & 150 & -- & 12 & Images & Simulator \\
HOMESAFEBENCH~\cite{gao2025homesafebench} & 12,900 & 5 & 12 & Images & Simulator \\
\midrule
\textbf{TSHA} & \textbf{66,668 QA pairs} & \textbf{12 parent / 64 fine} & \textbf{58 audited labels} & Images + Videos + Panoramas & Real + AIGC \\
\bottomrule
\end{tabular}}
\end{table*}

We evaluated 22 widely used Vision-Language Models (VLMs) on the TSHA benchmark's test set. Given that TSHA comprises both choice and QA questions, we use accuracy for choice questions and a weighted rubric for open-ended QA responses, as detailed later in \Cref{tab:qa_rubric}. The final score for each model is computed as the average of these two metrics. Our evaluation results in \Cref{tab:test_result} reveal that existing models exhibit substantial inadequacies in identifying indoor safety hazards. Notably, reinforcement learning (GRPO)-based fine-tuning of the Qwen2.5-VL~\cite{bai2025qwen2} model yields a performance improvement of up to +18.3 points, validating TSHA's efficacy as a training resource. Furthermore, evaluations on general benchmarks--including BLINK~\cite{fu2024blink}, AI2D~\cite{kembhavi2016diagram}, MMStar~\cite{chen2024rightwayevaluatinglarge}, MUIRBench~\cite{wang2024muirbenchcomprehensivebenchmarkrobust}, and SEEDBench2~\cite{li2023seedbench2benchmarkingmultimodallarge}--confirm that the model trained on TSHA achieves enhanced average scores, indicating that our benchmark not only boosts task-specific safety hazard assessment capabilities but also improves general visual reasoning performance.

Our contributions are summarized as follows:
\begin{itemize}
    \item We introduce TSHA, a comprehensive, realistic, and large-scale benchmark for indoor safety hazard assessment, with 66,668 validated QA pairs, 58 audited primary scene labels, and 12 parent / 64 fine-grained hazard types. Unlike prior works, it imposes no artificial restrictions on hazard categories, aiming to advance the hazard recognition capabilities of VLMs and establish a standardized training resource for both academic and industrial communities.
    \item We construct a rigorously curated test set incorporating Sora-generated video sequences and Hunyuan panoramic images with multiple embedded safety hazards. This suite is designed to enable comprehensive and stringent evaluation of model performance in identifying diverse and co-occurring hazardous situations.
    \item We conduct a systematic evaluation of 22 mainstream VLMs, exposing significant shortcomings in current models for safety hazard identification. Furthermore, by training baseline models on TSHA using reinforcement learning, we demonstrate a notable performance improvement on our TSHA test sets. Enhanced results on external benchmarks further confirm the generalization capability gained through TSHA-based training, underscoring its effectiveness as both an evaluation framework and a training resource.
\end{itemize}

\section{Related Work}
\label{sec:related_work}
\noindent\textbf{VLM for Safety.}
Visual-Language Models (VLMs) are increasingly being integrated into safety-critical applications, where their capacity to accurately interpret visual content, reason about real-world physical scenarios, and generate actionable insights is pivotal to preventing potential hazards and mitigating associated risks. This transformative potential has been most prominently realized in large-scale, public, and industrial domains, where reliable perception and reasoning capabilities directly impact operational safety and human well-being. For instance, in the field of autonomous driving, VLMs serve as a core component for robust situational awareness, enabling vehicles to perceive complex road environments, identify vulnerable road users, and navigate dynamic scenarios to avoid collisions~\cite{xu2024vlm,you2024v2x}. In industrial automation, VLMs power intelligent monitoring systems that continuously detect unsafe working conditions, equipment malfunctions, or operational anomalies, thereby ensuring worker safety in high-risk manufacturing and industrial environments~\cite{wu2025monitorvlm}. Furthermore, in disaster response scenarios, recent research explores leveraging VLMs to guide embodied agents in making real-time, safety-aware decisions within dynamically changing and hazardous environments, such as post-earthquake or fire-affected areas~\cite{zhou2024hazard}.

However, while these applications demonstrate the potential of VLMs in diverse scenarios, the nuanced and highly complex domain of private home safety remains significantly underexplored. Our work addresses this critical gap, focusing on adapting VLM capabilities for automated inspection of potential domestic hazards.

\noindent\textbf{Home Hazard Assessment Benchmarks.}
Existing benchmarks for home hazard detection adopt either text-based or vision-based methodologies. Text-based benchmarks, such as SafetyDetect~\cite{mullen2024don} and Safe-BeAl~\cite{huang2025framework}, pioneered this area by converting visual scenes into structured textual representations for analysis. While foundational, this text-centric approach creates an unavoidable information bottleneck. Specifically, nuanced visual details that are critical for accurate safety assessment are often lost or oversimplified during the image-to-text conversion process. This information loss directly impairs the ability to identify ambiguous or low-visibility hazards, which are particularly common in real-world home environments. Moreover, their scale remains limited, with SafetyDetect offering 1,000 samples across only 3 hazard categories.

Vision-based methods directly process visual inputs through VLMs. M-CoDAL~\cite{hassan2024coherence} pioneers this path with 2,000 Reddit-sourced images, while SafeAgentBench~\cite{yin2024safeagentbench} contributes 750 samples covering 10 hazard categories. The latest benchmark, HOMESAFEBENCH~\cite{gao2025homesafebench}, introduces 12,900 samples across 5 categories with interactive environment capabilities. To sum up, these approaches are still constrained by a narrow scope of predefined hazard categories, and large-scale studies rely on synthetic data, leading to critical gaps between simulation and reality.

In contrast to prior works, TSHA is fundamentally designed for greater realism and complexity. Unlike simulation-dependent work, TSHA prioritizes real-world data sources and constructs safety scenarios closer to everyday domestic environments, presenting significantly greater discrimination challenges for current models.

\section{The TSHA Benchmark}
\label{sec:tsha}
\begin{figure*}[t]
    \centering
    \includegraphics[width=\textwidth]{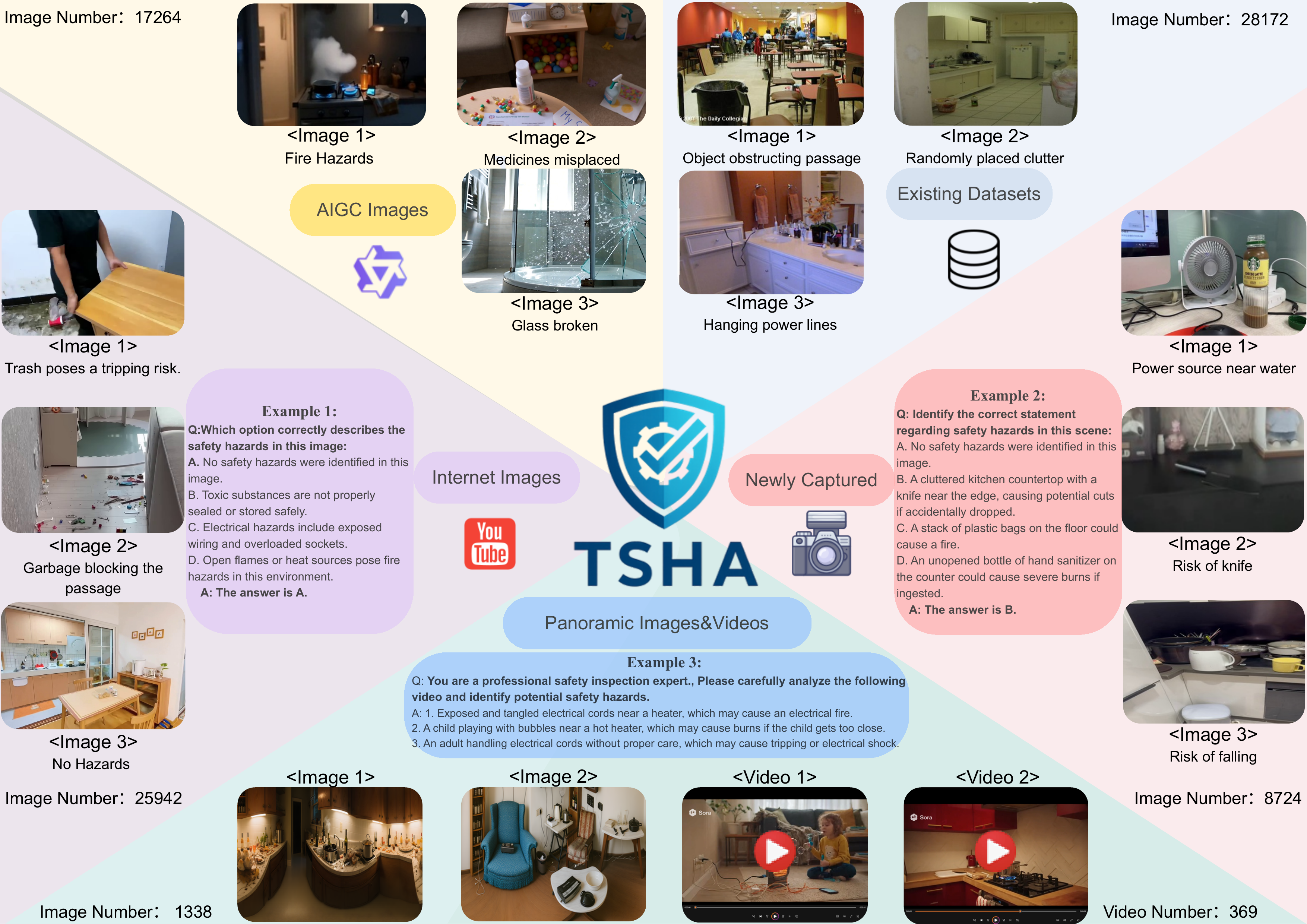}
    \caption{Overview of TSHA samples. TSHA combines existing datasets, internet images, AIGC images, newly captured images, videos, panoramas, and paired question-answer tasks to evaluate indoor safety hazard assessment.}
    \label{fig:datasets}
\end{figure*}

TSHA evaluates whether VLMs can identify indoor hazards from images, videos, and panoramic scenes. This section proceeds from data sources to the annotation pipeline, dataset statistics and hazard taxonomy, quality control and release filtering, and task formulation. Representative examples from existing datasets, internet images, AIGC images, newly captured images, videos, and panoramas are shown in \Cref{fig:datasets}.

\subsection{Data Sources}
\textbf{Training set.} TSHA training data come from complementary real and generated sources: existing datasets, internet images, newly captured images, and AIGC images. 

\textbf{Existing datasets} include NYU v2~\cite{silberman2012indoor} and MIT Indoor Scenes~\cite{quattoni2009recognizing}; the current validated annotation files provide 22,858 QA pairs from this source. These images supply broad indoor coverage and many negative or ordinary household scenes, which are necessary for preventing a benchmark from becoming a ``hazard always exists'' test.

\textbf{Internet images} are extracted from public daily-home videos after manually removing irrelevant prologues, epilogues, title cards, and non-indoor segments. The audited media inventory contains 60 usable source videos and 12,971 extracted frames, with 21,058 validated training QA pairs in the current release. We keep this source because real videos capture clutter, motion blur, camera viewpoint changes, and casual household layouts that are difficult to reproduce in clean simulator scenes. Since video frames can be temporally correlated, we report duplicate and near-duplicate statistics in \Cref{sec:quality_control} rather than assuming frame independence.

\textbf{AIGC images} are generated by closed-source text-to-video and text-to-image systems including Jimeng~\cite{gong2025seedream}, Hunyuan~\cite{sun2024hunyuan}, PixVerse, and HailuoAI, with prompts produced by Doubao~\cite{guo2025seed1}, Qwen~\cite{bai2025qwen2}, and DeepSeek~\cite{shao2024deepseekmath}; this source contributes 13,704 validated image QA pairs. AIGC is used as a supplement rather than the dominant source. Its role is to cover severe or rare hazards--such as open flame, electrical-water contact, structural collapse, gas/explosion risk, and dangerous child-access cases--that are unsafe, unethical, or impractical to stage in real homes. Because generated content can contain artifacts, AIGC samples are checked for physical plausibility and usability before release, as described in \Cref{sec:quality_control}.

\textbf{Newly captured images} are collected by 20 volunteers from everyday home and work scenes such as bedrooms, kitchens, living rooms, bathrooms, studies, and offices, contributing 7,034 validated QA pairs. Volunteers were asked to capture ordinary indoor environments rather than staged dramatic scenes, so this source increases realism and helps balance internet and AIGC biases. The current train manifest also contains 307 Hunyuan panorama open-ended QA pairs. In total, the validated training split contains 64,961 QA pairs.

\textbf{Test set.} The test set contains 1,707 QA pairs from six sources: existing datasets (392), newly captured images (102), internet images (296), AIGC images (210), Sora-generated videos (400), and Hunyuan panoramic images (307). The image subsets test static recognition under familiar and out-of-distribution sources. The Sora and panoramic subsets are intentionally reserved for evaluation because they stress capabilities that are not well measured by single-frame images: temporal evolution, wide fields of view, multiple simultaneous hazards, and the need to integrate evidence across a scene. This split design prevents the benchmark from measuring only memorization of a single data style.

\subsection{Annotation Pipeline}
\label{sec:annotation_pipeline}
\begin{figure*}[t]
    \centering
    \includegraphics[width=.99\textwidth]{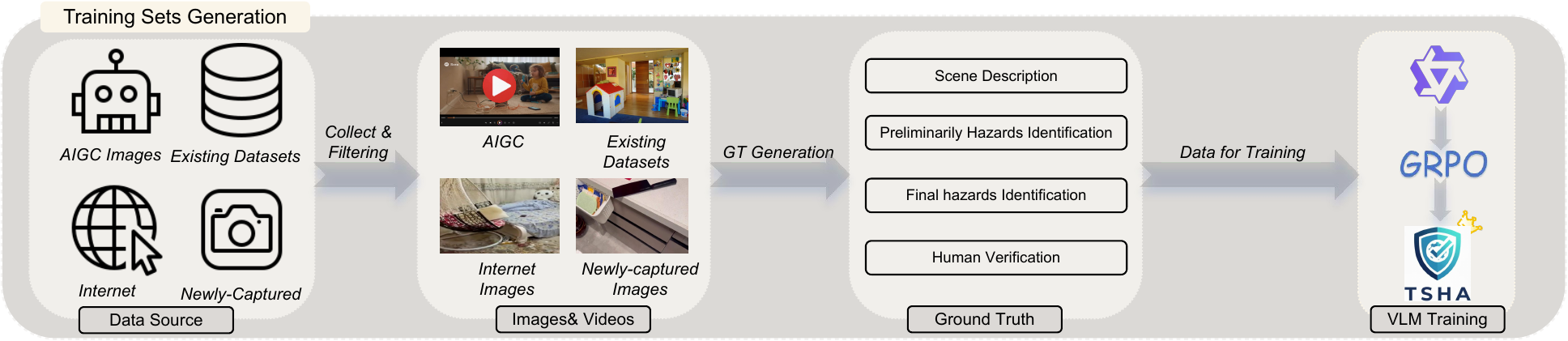}
    \caption{TSHA annotation pipeline. Each sample is processed through scene description, preliminary hazard identification, evidence-based hazard verification, self-correction, and human quality checks.}
    \label{fig:generation_pipeline}
\end{figure*}

The staged annotation and data-generation workflow is summarized in \Cref{fig:generation_pipeline}. We use GPT-4o~\cite{hurst2024gpt} as an automated annotation assistant because of its strong multimodal perception and stable instruction following. GPT-4o is not used as a one-pass labeler. The annotation pipeline is deliberately decomposed so that the model first observes, then hypothesizes, then verifies. This separation reduces the chance that an early safety judgment contaminates the visual description.

\textbf{Stage 1: objective scene description.} The model first describes visible objects, spatial relations, human presence, floor conditions, appliance states, and environmental cues without making a safety judgment. We ask for observable evidence rather than inferred intent. For example, a description may state that a cable lies across a walkway or that a pan is on an active stove, but it should not yet label the scene as hazardous.

\textbf{Stage 2: candidate hazard proposal.} The model then proposes candidate hazards across electrical, fire/heat, slip/trip/fall, obstruction/egress, falling-object, chemical/poison, child-safety, water/moisture, hygiene, and collision categories. At this stage, recall is prioritized: borderline candidates are allowed to remain if there is visible evidence, because later stages can reject weak candidates.

\textbf{Stage 3: evidence-based verification.} Each candidate hazard is checked against explicit visual evidence. The model must identify the object or condition that supports the hazard, the plausible consequence, and whether the risk is direct or speculative. Candidates are rejected when they rely only on generic prior knowledge, when the relevant object is not visible, or when the described configuration is ordinary and not safety-relevant. This stage is especially important for avoiding false positives such as treating every kitchen appliance, every cable, or every cluttered shelf as a hazard.

\textbf{Stage 4: QA construction and self-correction.} Verified hazards are converted into yes/no questions, single-choice questions, or open-ended hazard descriptions. For choice questions, distractors are generated from plausible but unsupported hazards so that the task measures visual grounding rather than keyword matching. A final self-correction pass checks that the answer key is consistent with the visual evidence, that negative examples are genuinely non-hazardous or low-risk, and that open-ended answers are concise and consequence-oriented.

\textbf{Stage 5: structural validation and human audit.} Generated files are then checked for media existence, decodability, question format, answer format, and path consistency. Human auditors check semantic correctness and AIGC physical plausibility as described in \Cref{sec:quality_control}. Rows marked as incorrect or unusable are removed or sent to adjudication. Thus, GPT-4o supplies scalable draft annotations, while the released benchmark is filtered through structural checks and human verification.

\subsection{Dataset Statistics and Hazard Taxonomy}
\label{sec:data_statistics}
\begin{figure*}[t]
\centering
\begin{minipage}[c]{0.245\textwidth}
    \centering
    \includegraphics[width=\linewidth]{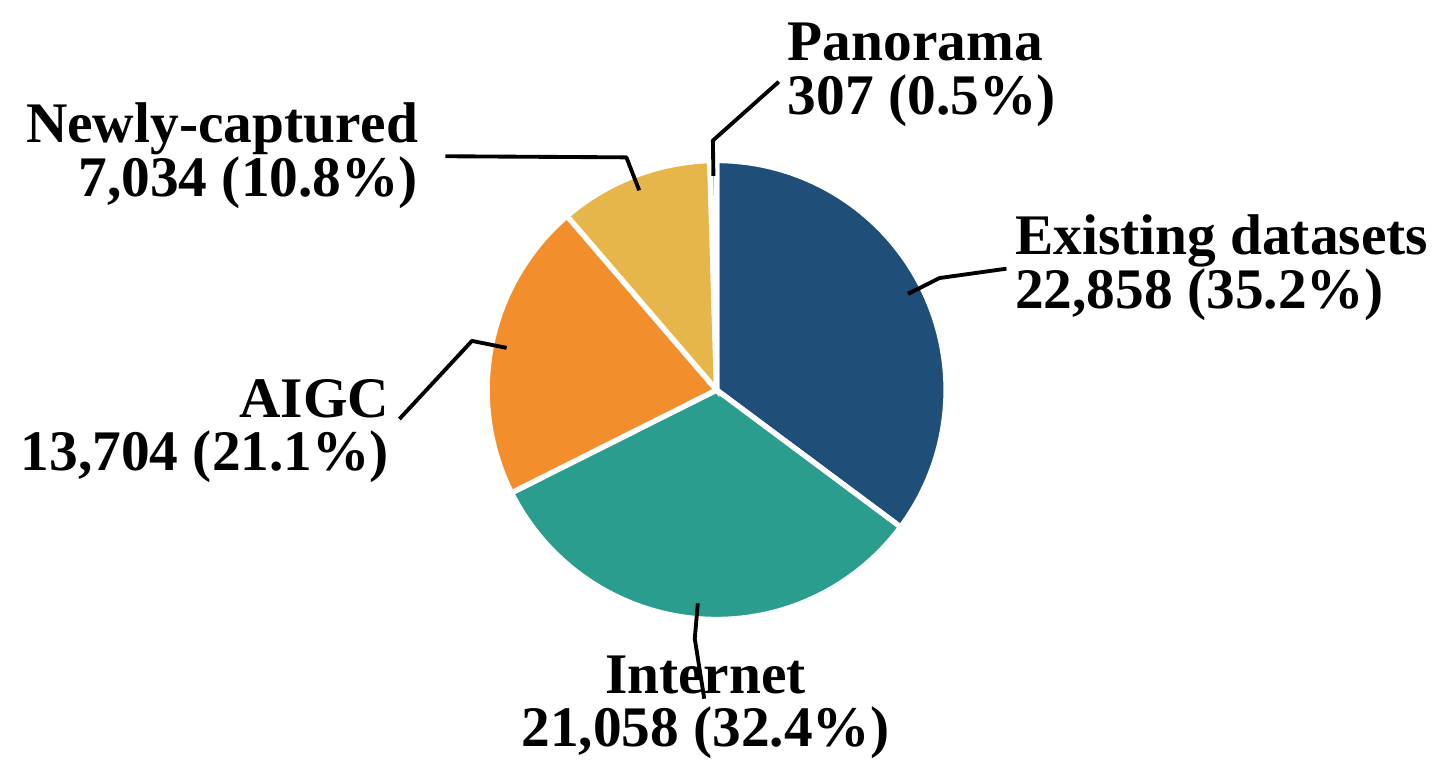}
\end{minipage}\hfill
\begin{minipage}[c]{0.235\textwidth}
    \centering
    \includegraphics[width=\linewidth]{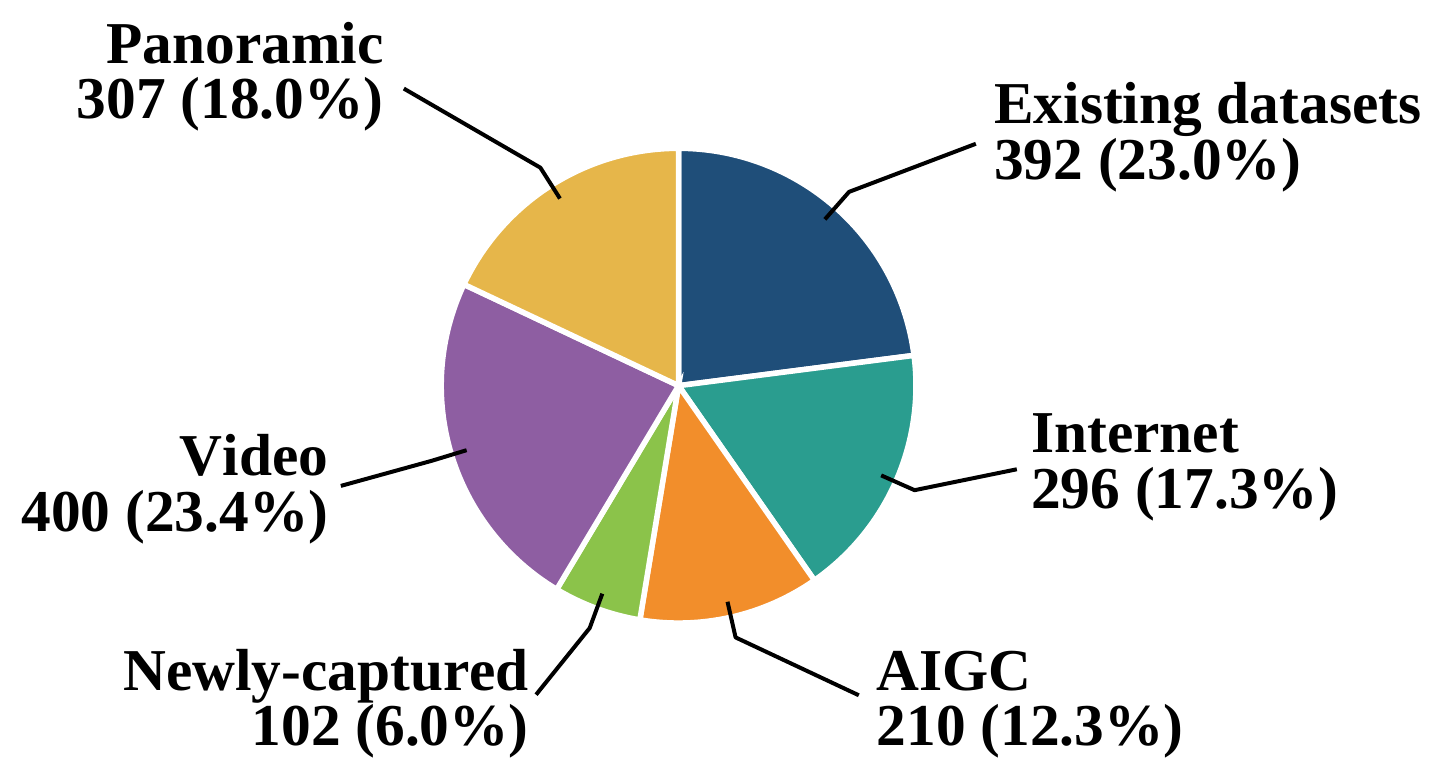}
\end{minipage}\hfill
\begin{minipage}[c]{0.505\textwidth}
    \centering
    \includegraphics[width=\linewidth]{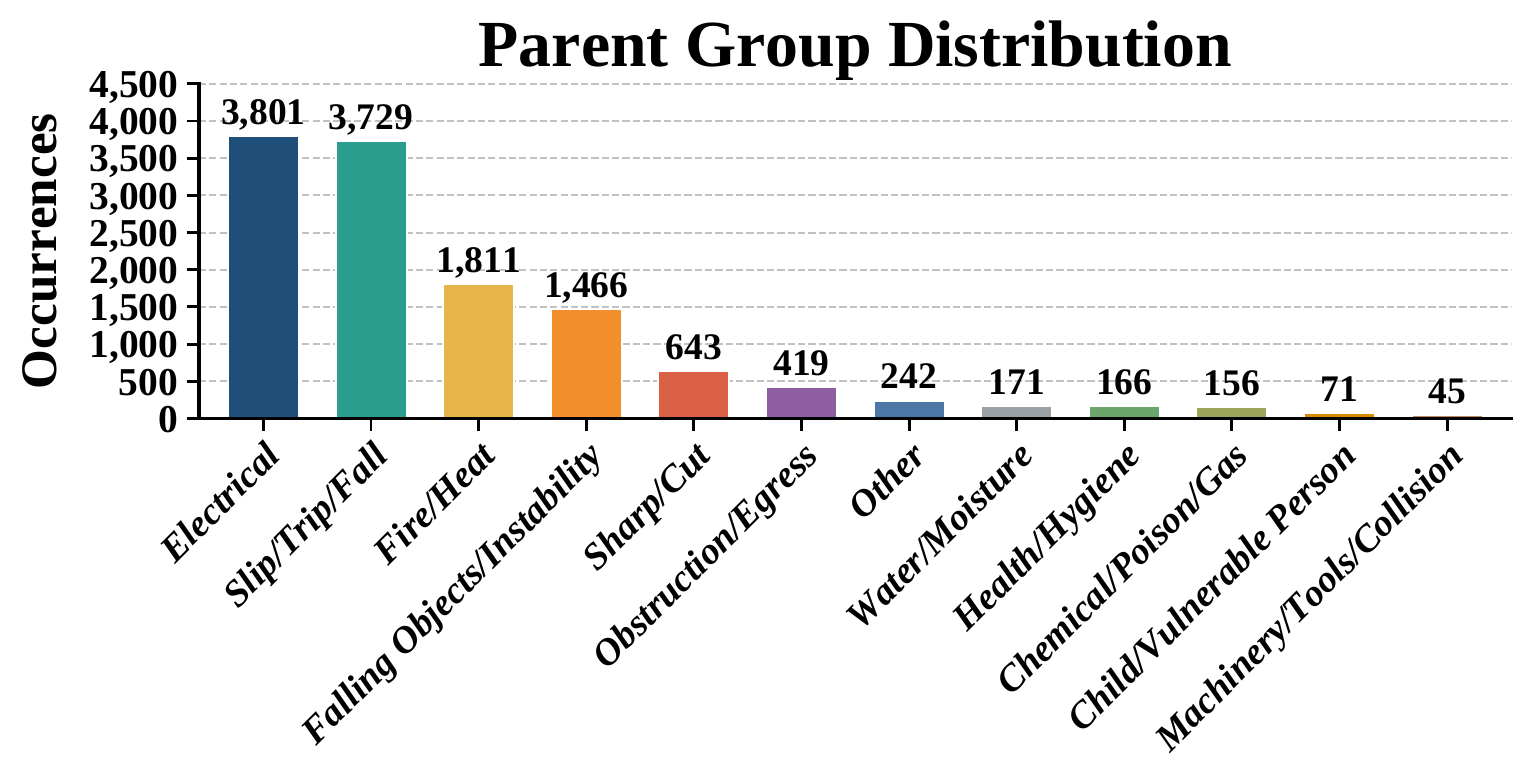}
\end{minipage}
\caption{QA-pair source distributions and parent hazard-group distribution in TSHA. Panels (a) and (b) show the training and test source distributions, respectively; panel (c) aggregates 12,720 normalized hazard occurrences into the 12 audited parent groups.}
\label{fig:data_distribution}
\end{figure*}

\Cref{tab:dataset_stats} and \Cref{fig:data_distribution} summarizes the current validated data scale. To avoid ambiguity, we distinguish media items from QA pairs. Most image-based training items are paired with two QA pairs, while panorama/video items are paired with one open-ended QA item.

\begin{table*}[t]
\centering
\caption{TSHA data statistics. Counts are reported as QA pairs unless media count is explicitly shown.}
\label{tab:dataset_stats}
\tabfont
\setlength{\tabcolsep}{17pt}
\renewcommand{\arraystretch}{1.05}{%
\begin{tabular}{lrrrrr}
\toprule
\textbf{Split / source} & \textbf{Media items} & \textbf{Yes/No QA} & \textbf{Multiple-choice QA} & \textbf{Open QA} & \textbf{Total QA} \\
\midrule
Train: existing indoor datasets & 11,429 & 11,429 & 11,429 & 0 & 22,858 \\
Train: internet frames & 10,529 & 10,529 & 10,529 & 0 & 21,058 \\
Train: AIGC images & 6,852 & 6,852 & 6,852 & 0 & 13,704 \\
Train: newly captured images & 3,517 & 3,517 & 3,517 & 0 & 7,034 \\
Train: Hunyuan panoramas & 307 & 0 & 0 & 307 & 307 \\
\midrule
\textbf{Training total} & \textbf{32,634} & \textbf{32,327} & \textbf{32,327} & \textbf{307} & \textbf{64,961} \\
\midrule
Test: existing indoor datasets & 196 & 196 & 196 & 0 & 392 \\
Test: internet images & 148 & 148 & 148 & 0 & 296 \\
Test: AIGC images & 105 & 105 & 105 & 0 & 210 \\
Test: newly captured images & 51 & 51 & 51 & 0 & 102 \\
Test: Sora videos & 400 & 0 & 0 & 400 & 400 \\
Test: Hunyuan panoramas & 307 & 0 & 0 & 307 & 307 \\
\midrule
\textbf{Test total} & \textbf{1,207} & \textbf{500} & \textbf{500} & \textbf{707} & \textbf{1,707} \\
\midrule
\textbf{Dataset total} & \textbf{33,841} & \textbf{32,827} & \textbf{32,827} & \textbf{1,014} & \textbf{66,668} \\
\bottomrule
\end{tabular}}
\end{table*}

The source-media inventory contains 40,051 media items covering 58 primary scene labels, from which we construct the current validated annotation release. The largest scene groups are bedroom (7,959 media items), living room (6,724), kitchen (4,950), bathroom (2,242), laundry room (1,358), corridor (1,084), dining room (1,016), gym (885), closet (836), office (764), and playroom (741). This distribution is intentionally long-tailed. Rare but safety-relevant scenes, including balconies, garages, laboratories, staircases, stores, and storage rooms, appear less often, but they help test whether models remain reliable beyond common home environments.

The hazard taxonomy includes 12 parent groups and 64 mutually exclusive fine-grained hazard types, built from 12,720 normalized hazard occurrences and 12,487 unique normalized hazard phrases. As shown in \Cref{fig:data_distribution}, the largest parent groups are Electrical (3,801; 29.9\%), Slip/Trip/Fall (3,729; 29.3\%), Fire/Heat (1,811; 14.2\%), and Falling Objects/Instability (1,466; 11.5\%). The resulting distribution is imbalanced, as expected in indoor safety data: common hazards dominate the corpus, while severe or rare hazards are supplemented with AIGC and panoramic/video test cases.



\textbf{Hazard taxonomy construction.}
\label{sec:hazard_taxonomy}
The taxonomy is built from normalized hazard phrases. We first collect hazard descriptions from annotation and auditing, normalize lexical variants, and group phrases according to the underlying physical risk. Fine-grained labels are merged only when they describe the same actionable hazard. For example, ``wet floor'', ``water on the floor'', and ``slippery tile'' are merged into a slippery-surface type, while ``loose electrical cords'', ``overloaded power strip'', and ``electricity near water'' remain separate because they call for different mitigation actions. This rule keeps near-duplicate phrases from inflating the taxonomy while preserving distinctions that are important for safety assessment.

Each fine-grained type is assigned to exactly one parent group. Parent groups support analysis and reporting, whereas fine-grained types retain hazard diversity. Borderline cases are assigned by the primary injury or failure mechanism. For example, a cable across a walkway is categorized as an electrical-cord trip hazard under the Electrical parent group when the textual evidence focuses on the cable, while generic floor clutter belongs to Slip/Trip/Fall. A shelf leaning over a bed is assigned to Falling Objects/Instability, and a blocked doorway is assigned to Obstruction/Egress. This mutually exclusive assignment keeps the statistics interpretable and prevents the same phrase from being counted under multiple parent groups. As a result, \Cref{fig:data_distribution}(c) and \Cref{tab:hazard_taxonomy_parent} report the same parent-group distribution, while \Cref{tab:hazard_taxonomy_parent} also lists representative fine-grained types for each parent group.

\begin{table*}[t]
\centering
\caption{Audited parent hazard taxonomy. Parent occurrences sum the mutually exclusive fine-grained types under each group.}
\label{tab:hazard_taxonomy_parent}
\tabfont
\setlength{\tabcolsep}{4pt}
\renewcommand{\arraystretch}{1.05}{%
\begin{tabular}{lrrp{0.62\textwidth}}
\toprule
\textbf{Parent group} & \textbf{Fine types} & \textbf{Occurrences} & \textbf{Representative fine-grained types} \\
\midrule
Electrical & 7 & 3,801 & Loose electrical cords (1,404); electricity near water (960); overloaded outlets or power strips (643); sparking/short-circuit/overheating (268); exposed or damaged wiring (250) \\
Slip/Trip/Fall & 9 & 3,729 & Floor clutter or low obstacle (2,079); wet or slippery walking surface (652); loose rug, mat, or fabric (365); fall from height or unguarded edge (240); general slip/trip/fall risk (114) \\
Fire/Heat & 8 & 1,811 & Unattended cooking/stove/oven (1,153); flammables near heat (252); gas leak or explosion risk (146); open flame or active fire (101); hot surface, hot liquid, burn, or scald (66) \\
Falling Objects/Instability & 7 & 1,466 & Furniture tip-over or collapse (508); structural damage or collapse (432); open window or balcony guardrail issue (193); unstable stacking or storage (162); object placed near edge (119) \\
Sharp/Cut & 4 & 643 & Knives or exposed blades (258); broken glass or sharp mirror/glass edges (250); general cut/laceration risk (78); sharp tools, needles, or scissors (57) \\
Obstruction/Egress & 4 & 419 & Blocked emergency exit or escape route (292); blocked doorway or walking path (64); general obstruction or crowding (53); obstructed visibility, warning, or signage (10) \\
Water/Moisture & 3 & 171 & Pipe, faucet, or ceiling leak (87); moisture or water-damage risk (45); standing water or flooding (39) \\
Health/Hygiene & 5 & 166 & Mold or damp-health risk (78); waste, pests, or hygiene issue (61); biohazard, dirty water, or contamination (12); poor air quality or ventilation (9); general health/hygiene risk (6) \\
Chemical/Poison/Gas & 4 & 156 & Medication or small-item ingestion (65); gas, carbon-monoxide, or toxic-fume risk (56); cleaning chemical exposure or spill (34); general chemical/poison risk (1) \\
Child/Vulnerable Person & 5 & 71 & Falling object near child/crib (21); child choking, strangulation, or entanglement (18); supervision risk (16); child fall from window/crib/elevated area (15); child access to dangerous object/substance (1) \\
Machinery/Tools/Collision & 6 & 45 & Poor visibility or glare (14); head-bump/protrusion risk (12); general machinery/tool/collision risk (8); machinery/tool guarding or pinch risk (6); robotic or moving-device collision (3) \\
Other & 2 & 242 & Low-specificity or ambiguous hazard (237); property-only food spoilage/energy-waste issue (5) \\
\bottomrule
\end{tabular}}
\end{table*}


\subsection{Quality Control and Filtering}
\label{sec:quality_control}

\begin{table*}[t]
\centering
\caption{Human-audit and physical-plausibility checks.}
\label{tab:quality_audit}
\tabfont
\setlength{\tabcolsep}{17pt}
\renewcommand{\arraystretch}{1.05}{%
\begin{tabular}{lccc}
\toprule
\textbf{Audit item} & \textbf{Auditor 1} & \textbf{Auditor 2} & \textbf{Agreement / notes} \\
\midrule
Semantic label inconsistency rate & 1.10\% & 4.26\% & 95.13\% exact agreement \\
Weighted semantic mismatch rate & 0.41\% & 2.95\% & 41 rows sent to adjudication \\
AIGC image physical-plausibility pass rate & 98.50\% & 99.00\% & 98.50\% exact agreement \\
AIGC image usable-for-evaluation rate & 98.75\% & 99.00\% & 98.75\% exact agreement; invalid frames discarded \\
Sora video physical-plausibility pass rate & 97.25\% & 99.50\% & 96.75\% exact agreement \\
Sora video usable-for-evaluation rate & 97.25\% & 99.50\% & 96.75\% exact agreement; unusable videos removed \\
\bottomrule
\end{tabular}}
\end{table*}

\textbf{Structural-format check.} We first check media accessibility, file decodability, path consistency, modality consistency, question format, answer format, and split assignment. Samples with broken media, malformed QA fields, inconsistent paths, or mismatched modality labels are removed before semantic or physical audits.

\textbf{Semantic-label and error check.} For annotation quality, we audit a stratified 10\% sample. The sample is stratified by split, data source, and question type, so rare but important sources such as Sora videos and Hunyuan panoramas are not overwhelmed by common image sources. Each auditor sees the media, question, answer key, and annotation text, and checks whether the answer is supported by visible evidence. As shown in Table~IV, the semantic label inconsistency rates are 1.10\% and 4.26\% for the two auditors, with 95.13\% exact agreement. The weighted semantic mismatch rates are 0.41\% and 2.95\%. The main error types are false hazards, wrong yes/no labels, missed hazards in selected options, and ambiguous borderline cases.

\textbf{Generated-media plausibility and usability check.} For generated media, we check whether AIGC images and Sora videos are physically coherent and usable for safety reasoning, rather than judging only whether the hazard label is correct. The AIGC audit uses 10\% media-level sample stratified across split and generation source, while the Sora audit covers 100\% of the Sora test subset. Blocking failures include collage-like compositions, severe tearing artifacts, unviewable media, impossible object geometry, and temporal artifacts that make the hazard uninterpretable. In Table~IV, AIGC image physical-plausibility pass rates are 98.50\% and 99.00\%, and usable-for-evaluation rates are 98.75\% and 99.00\%. For Sora videos, physical-plausibility and usable-for-evaluation pass rates are 97.25\% and 99.50\%. Samples judged unusable by either auditor are removed or sent to adjudication.

\textbf{Internet-source diversity and privacy check.} Because internet images are extracted from videos, we audit both source diversity and frame redundancy. Frames are grouped by source video before analysis, preventing a long video from being treated as many independent scenes. We then compute perceptual-hash similarity to estimate near-duplicate rates. Exact duplicates are negligible at 0.008\%, while adjacent-frame near-duplicate rates range from 8.16\% to 33.10\% under increasingly relaxed dHash thresholds. These results show expected temporal correlation but not domination by exact duplicates. For release, internet-derived media are also filtered for faces, readable private text, addresses, license plates, personal documents, and other sensitive identifiers. Sensitive samples are redacted, removed, or released only as metadata when redistribution is not appropriate.
\begin{table*}[t]
\centering
\caption{Open-ended QA scoring rubric. Accuracy is weighted most heavily because missed or hallucinated hazards directly affect safety.}
\label{tab:qa_rubric}
\tabfont
\setlength{\tabcolsep}{10pt}
\renewcommand{\arraystretch}{1.05}{%
\begin{tabular}{lp{0.36\textwidth}p{0.36\textwidth}c}
\toprule
\textbf{Dimension} & \textbf{High score} & \textbf{Low score} & \textbf{Weight} \\
\midrule
Accuracy & Identifies the major visible hazards, grounds each hazard in observable evidence, and gives a plausible consequence. & Misses major hazards, invents unsupported hazards, confuses benign objects with hazards, or gives the wrong consequence. & 0.7 \\
Conciseness & Provides a short, actionable list of hazards without redundant background description. & Produces long scene captions, repeated warnings, or irrelevant advice that hides the actual risk. & 0.2 \\
Coherence & Uses clear causal language linking condition and consequence, e.g., ``water near outlet may cause electric shock.'' & Gives fragmented, contradictory, or hard-to-interpret statements. & 0.1 \\
\bottomrule
\end{tabular}}
\end{table*}
\subsection{Task Formulation}
\label{sec:task_formulation}
TSHA uses three complementary task formats. \textbf{Yes/no questions} ask whether a scene contains any safety hazard. They test conservative hazard presence detection and expose false-positive tendencies in ordinary indoor scenes. \textbf{Single-choice questions} present four candidate assessments, only one of which is supported by the image or video. Distractors are designed to be plausible hazard descriptions, so a model must ground its answer in visible evidence rather than select any safety-related phrase. \textbf{Open-ended questions} ask the model to identify the most significant visible hazards and describe their consequences without predefined options.

The three formats stress different capabilities. Yes/no questions measure binary risk recognition. Single-choice questions measure discriminative grounding among plausible alternatives. Open-ended questions measure free-form hazard discovery, evidence selection, and concise explanation. We include open-ended questions because real safety inspection is not a closed-set classification problem: a useful assistant should identify multiple hazards, ignore unsupported risks, and explain why the scene is dangerous.

For open-ended video and panorama cases, the expected output is a short list of one to three hazards, each written as a concrete risk and consequence. We reward responses that identify visible hazards and explain their likely harm; we penalize hallucinated hazards, vague warnings without visual evidence, excessive verbosity, and missing major hazards. This format is intentionally stricter than generic captioning because a safety system should not merely describe the room--it should surface actionable risks.

\begin{table*}[t]
\centering
\caption{Evaluation on TSHA. For QA tasks, we report answer accuracy (QA acc.), conciseness (QA con.), coherence (QA coh.), and weighted QA overall. For choice questions, T/F and single-choice accuracies are averaged as CQ overall. Final Avg. is the mean of QA overall and CQ overall.}
\label{tab:test_result}
\tabfont
\setlength{\tabcolsep}{9pt}
\renewcommand{\arraystretch}{1.05}{%
\begin{tabular}{l c| c c c c |c c c|c}
\toprule
\textbf{Model} & \textbf{Param.} & \multicolumn{4}{c|}{\textit{QA Questions}} & \multicolumn{3}{c|}{\textit{Choice Questions}} & \textbf{Avg.} \\
&& QA acc. & QA con. & QA coh. & QA overall & T/F acc. & SC acc. & CQ overall & \\
\midrule
\multicolumn{10}{c}{\textit{Closed-source VLMs}} \\
\midrule
Gemini-2.5-Flash~\cite{comanici2025gemini} & -- & 75.3 & 80.1 & 84.0 & 77.1 & 37.4 & 54.2 & 45.8 & 61.5 \\
Gemini-2.5-Pro~\cite{comanici2025gemini} & -- & 76.3 & 81.5 & 85.4 & 78.3 & 33.9 & 55.0 & 44.4 & 61.4 \\
Claude-3.7-Sonnet~\cite{claude} & -- & 71.1 & 79.9 & 82.1 & 74.0 & 82.7 & 91.4 & 87.1 & 80.6 \\
Claude-4-Sonnet~\cite{claude} & -- & 76.7 & 80.3 & 85.5 & 78.8 & 75.6 & 86.8 & 81.2 & 80.0 \\
\midrule
\multicolumn{10}{c}{\textit{Open-source VLMs}} \\
\midrule
Gemma3~\cite{team2025gemma} & 4B & 35.2 & 39.5 & 41.5 & 36.7 & 32.1 & 28.3 & 30.2 & 33.4 \\
Gemma3~\cite{team2025gemma} & 12B & 34.9 & 39.3 & 40.7 & 36.4 & 30.6 & 35.3 & 33.0 & 34.7 \\
Gemma3~\cite{team2025gemma} & 27B & 36.5 & 40.1 & 41.3 & 37.7 & 27.3 & 56.3 & 41.8 & 39.8 \\
InternVL2~\cite{chen2024internvl} & 26B & 69.1 & 78.5 & 79.8 & 72.1 & 79.5 & 85.3 & 82.4 & 77.3 \\
InternVL2.5~\cite{chen2024expanding} & 8B & 68.5 & 77.5 & 79.1 & 71.4 & 80.6 & 87.6 & 84.1 & 77.8 \\
InternVL2.5-MPO~\cite{chen2024expanding} & 8B & 70.5 & 70.5 & 81.0 & 71.6 & 66.2 & 89.0 & 77.6 & 74.6 \\
InternVL2.5~\cite{chen2024expanding} & 26B & 70.2 & 80.1 & 80.8 & 73.2 & 85.3 & 91.3 & 88.3 & 80.8 \\
InternVL3~\cite{zhu2025internvl3} & 8B & 66.1 & 74.4 & 77.9 & 68.9 & 90.6 & 92.0 & 91.3 & 80.1 \\
Mimo-RL~\cite{coreteam2025mimovltechnicalreport} & 7B & 72.0 & 60.9 & 62.7 & 68.9 & 86.2 & 74.8 & 80.5 & 74.7 \\
Mimo-SFT~\cite{coreteam2025mimovltechnicalreport} & 7B & 75.7 & 72.9 & 76.2 & 75.2 & 86.3 & 84.7 & 85.5 & 80.4 \\
MiniCPM~\cite{minicpm4} & 4B & 65.6 & 77.2 & 77.1 & 69.1 & 85.1 & 89.0 & 87.1 & 78.1 \\
Mistral-Small-3.1~\cite{rastogi2025magistral} & 24B & 18.8 & 20.6 & 21.1 & 19.4 & 31.3 & 64.2 & 47.8 & 33.6 \\
Ovis2.5~\cite{lu2025ovis25technicalreport} & 2B & 72.7 & 68.1 & 72.1 & 71.7 & 84.1 & 73.2 & 78.7 & 75.2 \\
Ovis2.5~\cite{lu2025ovis25technicalreport} & 9B & 75.8 & 65.5 & 68.1 & 73.0 & 81.3 & 80.6 & 81.0 & 77.0 \\
\midrule
Ovis2~\cite{lu2024ovis} & 1B & 58.8 & 59.3 & 64.8 & 59.5 & 76.6 & 72.0 & 74.3 & 66.9 \\
\rowcolor{lightgrayrow}Ovis2 + TSHA & 1B & 64.0 & 75.4 & 77.5 & 67.6 & 88.6 & 98.8 & 93.7 & 80.7 \\
\rowcolor{lightbluerow}$\Delta$ & 1B & +5.2 & +16.1 & +12.7 & +8.1 & +12.0 & +26.8 & +19.4 & +13.8 \\
\midrule
MiniCPM~\cite{minicpm4} & 4B & 65.6 & 77.2 & 77.1 & 69.1 & 85.1 & 89.0 & 87.1 & 78.1 \\
\rowcolor{lightgrayrow}MiniCPM + TSHA & 4B & 69.3 & 79.6 & 82.3 & 72.7 & 89.1 & 99.8 & 94.5 & 83.6 \\
\rowcolor{lightbluerow}$\Delta$ & 4B & +3.7 & +2.4 & +5.2 & +3.6 & +4.0 & +10.8 & +7.4 & +5.5 \\
\midrule
Qwen2.5-VL~\cite{bai2025qwen2} & 3B & 57.2 & 70.2 & 70.3 & 61.1 & 71.4 & 63.0 & 67.2 & 64.2 \\
\rowcolor{lightgrayrow}Qwen2.5-VL + TSHA & 3B & 68.2 & 78.6 & 80.2 & 71.5 & 90.8 & 96.0 & 93.4 & 82.5 \\
\rowcolor{lightbluerow}$\Delta$ & 3B & +11.0 & +8.4 & +9.9 & +10.4 & +19.4 & +33.0 & +26.2 & +18.3 \\
Qwen2.5-VL~\cite{bai2025qwen2} & 7B & 68.2 & 77.1 & 79.5 & 71.1 & 86.3 & 77.3 & 81.8 & 76.5 \\
\rowcolor{lightgrayrow}Qwen2.5-VL + TSHA & 7B & 71.6 & 80.2 & 81.8 & 74.3 & 90.3 & 96.8 & 93.6 & 84.0 \\
\rowcolor{lightbluerow}$\Delta$ & 7B & +3.4 & +3.1 & +2.3 & +3.2 & +3.0 & +19.5 & +11.8 & +7.5 \\
Qwen2.5-VL~\cite{bai2025qwen2} & 32B & 73.6 & 79.5 & 82.5 & 75.7 & 86.6 & 81.3 & 84.0 & 79.9 \\
\rowcolor{lightgrayrow}Qwen2.5-VL + TSHA & 32B & 75.5 & 80.2 & 83.7 & 77.3 & 90.8 & 96.0 & 93.4 & 85.4 \\
\rowcolor{lightbluerow}$\Delta$ & 32B & +1.9 & +0.7 & +1.2 & +1.6 & +3.2 & +14.7 & +8.6 & +5.5 \\
\bottomrule
\end{tabular}}
\end{table*}
\section{Experiments}
\label{sec:experiments}
\subsection{Experimental Setup}
We evaluate 22 VLMs, including both closed-source and open-source systems. The model pool covers commercial systems, medium-to-large open-source models, and smaller models that are realistic candidates for local deployment. This range is important because safety monitoring is not limited to the strongest cloud models: some applications require lower-latency or privacy-preserving local models.

To test whether TSHA can serve as a training resource, we fine-tune Qwen2.5-VL-3B, Qwen2.5-VL-7B, Qwen2.5-VL-32B, Ovis2-1B, and MiniCPM-4B using GRPO~\cite{shao2024deepseekmath} with the ms-swift framework~\cite{zhao2024swiftascalablelightweightinfrastructure}; for general-benchmark transfer, we additionally report Qwen2.5-VL-72B results where available. Training experiments are conducted on 8 NVIDIA H200-141G GPUs. For each base model, the comparison is between the original model and the same model after TSHA training, so improvements are attributable to the safety-hazard supervision rather than to changing the backbone family.

We evaluate transfer to BLINK~\cite{fu2024blink}, AI2D~\cite{kembhavi2016diagram}, MMStar~\cite{chen2024rightwayevaluatinglarge}, MUIRBench~\cite{wang2024muirbenchcomprehensivebenchmarkrobust}, and SEEDBench2~\cite{li2023seedbench2benchmarkingmultimodallarge} using VLMEvalKit~\cite{duan2024vlmevalkit}. These benchmarks are not safety-specific, so they test whether safety fine-tuning damages or preserves general multimodal reasoning. We report per-benchmark results rather than only averages, because average transfer can hide small regressions on individual datasets.

For choice questions, we use accuracy. For open-ended QA, we score responses along three dimensions: accuracy, conciseness, and coherence. The primary QA score is
\begin{equation}
S_{\mathrm{QA}} = 0.7S_{\mathrm{accuracy}} + 0.2S_{\mathrm{conciseness}} + 0.1S_{\mathrm{coherence}},
\end{equation}
where accuracy receives the highest weight because missed or hallucinated hazards directly affect safety. The final score averages open-ended QA and choice-question performance:
\begin{equation}
S_{\mathrm{overall}} = \frac{S_{\mathrm{QA}} + S_{\mathrm{CQ}}}{2}.
\end{equation}
The 0.7/0.2/0.1 weighting reflects the safety-critical nature of the task. Missing or hallucinating a hazard is more serious than a stylistic issue, so accuracy receives the largest weight. Conciseness is still included because verbose answers can obscure the actionable risk. Coherence is included because a response should connect the visible condition to a plausible consequence. The complete scoring rubric is summarized in \Cref{tab:qa_rubric}.

Because LLM judges can overestimate improvements, we report multi-evaluator and human-evaluation results in \Cref{tab:different_eval}. The main TSHA leaderboard uses GPT-4o only as the scalable automatic evaluator, while GPT-4o itself is not reported as a tested model in \Cref{tab:test_result}. We treat human evaluation as the conservative estimate of practical improvement. When evaluator scores differ, we emphasize the direction of improvement agreed on by multiple evaluators rather than the exact absolute score of a single judge.

\subsection{Main Results on TSHA}
\Cref{tab:test_result} shows that TSHA is challenging for current VLMs. Performance varies widely across both closed-source and open-source models: Claude-3.7-Sonnet, Claude-4-Sonnet, and several strong open-source VLMs perform well, while smaller or less safety-specialized models remain far behind. Claude-4-Sonnet is evaluated with the same valid-response scoring protocol as the other closed-source models, making the comparison consistent.

Training on TSHA improves every fine-tuned model on the TSHA test set. Qwen2.5-VL-3B improves by 18.3 points, Ovis2-1B by 13.8 points, Qwen2.5-VL-7B by 7.5 points, MiniCPM-4B by 5.5 points, and Qwen2.5-VL-32B by 5.5 points. These gains are especially large for choice questions, but open-ended QA also improves, suggesting that TSHA provides useful hazard-specific supervision rather than only teaching option selection.

The largest gains occur for smaller and mid-sized models, which suggests that TSHA supplies missing safety-specific concepts rather than merely reinforcing capabilities already present in the strongest models. For example, Qwen2.5-VL-3B improves by +10.4 on open-ended QA overall and +26.2 on choice-question overall. This pattern indicates that the model learns both free-form hazard description and discriminative grounding among plausible options. Larger models start from stronger general visual reasoning, so their absolute gains are smaller, but Qwen2.5-VL-32B still improves by +1.6 on open-ended QA and +8.6 on choice questions.

The results also show that open-ended safety assessment remains difficult. Several strong models achieve high single-choice accuracy but lower open-ended QA scores, indicating that selecting among options is easier than discovering and explaining hazards without choices. This distinction is central to TSHA: a deployment-oriented safety assistant cannot rely on a pre-written list of options, and it must avoid both missing real hazards and hallucinating unsupported ones.

\begin{table*}[t]
\centering
\caption{Evaluation on general benchmarks. TSHA improves average performance for 3B, 7B, and 72B Qwen2.5-VL models, but the 72B model shows small drops on Blink and MMStar, indicating a mild safety-specialization trade-off.}
\label{tab:other_bench}
\tabfont
\setlength{\tabcolsep}{17pt}
\renewcommand{\arraystretch}{1.05}{%
\begin{tabular}{lccccccc}
\toprule
\textbf{Model} & \textbf{Size} & \textbf{BLINK} & \textbf{AI2D} & \textbf{MMStar} & \textbf{MUIR} & \textbf{SEED} & \textbf{Average} \\
\midrule
Qwen2.5-VL~\cite{bai2025qwen2} & 3B & 47.6 & 81.6 & 55.9 & 47.7 & 67.6 & 60.1 \\
\rowcolor{lightgrayrow}Qwen2.5-VL + TSHA & 3B & 49.6 & 82.7 & 57.1 & 49.6 & 69.3 & 61.7 \\
\rowcolor{lightbluerow}$\Delta$ & 3B & +2.0 & +1.1 & +1.2 & +1.9 & +1.7 & +1.6 \\
\midrule
Qwen2.5-VL~\cite{bai2025qwen2} & 7B & 56.4 & 83.9 & 63.9 & 59.6 & 70.4 & 66.8 \\
\rowcolor{lightgrayrow}Qwen2.5-VL + TSHA & 7B & 56.9 & 85.6 & 64.3 & 61.9 & 71.2 & 68.0 \\
\rowcolor{lightbluerow}$\Delta$ & 7B & +0.5 & +1.5 & +0.4 & +2.3 & +0.8 & +1.2 \\
\midrule
Qwen2.5-VL~\cite{bai2025qwen2} & 72B & 64.4 & 88.7 & 70.8 & 70.7 & 73.0 & 73.5 \\
\rowcolor{lightgrayrow}Qwen2.5-VL + TSHA & 72B & 62.2 & 89.6 & 69.7 & 73.7 & 74.2 & 73.9 \\
\rowcolor{lightbluerow}$\Delta$ & 72B & -2.2 & +0.9 & -1.1 & +3.0 & +1.2 & +0.4 \\
\bottomrule
\end{tabular}}
\end{table*}

\begin{table*}[t]
\centering
\caption{Ablation over TSHA training sources on general benchmarks. Each source contributes positive average transfer; combining all sources gives the best average result.}
\label{tab:ablation}
\tabfont
\setlength{\tabcolsep}{9.5pt}
\renewcommand{\arraystretch}{1.05}{%
\begin{tabular}{lcccc|ccccc|c}
\toprule
\textbf{Model} & \textbf{AIGC} & \textbf{Existing} & \textbf{Internet} & \textbf{New} & \textbf{BLINK} & \textbf{AI2D} & \textbf{MMStar} & \textbf{MUIR} & \textbf{SEED} & \textbf{Avg.} \\
\midrule
Qwen2.5-VL-3B~\cite{bai2025qwen2} & & & & & 47.6 & 81.6 & 55.9 & 47.7 & 67.6 & 60.1 \\
\midrule
Qwen2.5-VL-3B & \cmark & & & & 49.1 & 82.3 & 56.9 & 48.9 & 68.9 & 61.2 \\
Qwen2.5-VL-3B & & \cmark & & & 49.1 & 82.5 & 57.1 & 48.1 & 69.0 & 61.2 \\
Qwen2.5-VL-3B & & & \cmark & & 48.9 & 82.4 & 56.7 & 49.5 & 69.3 & 61.4 \\
Qwen2.5-VL-3B & & & & \cmark & 49.2 & 82.4 & 57.1 & 49.3 & 68.6 & 61.3 \\
\rowcolor{lightgrayrow}Qwen2.5-VL-3B & \cmark & \cmark & \cmark & \cmark & 49.6 & 82.7 & 57.1 & 49.6 & 69.3 & 61.7 \\
\midrule
Qwen2.5-VL-7B~\cite{bai2025qwen2} & & & & & 56.4 & 83.9 & 63.9 & 59.6 & 70.4 & 66.8 \\
\midrule
Qwen2.5-VL-7B & \cmark & & & & 56.9 & 85.4 & 64.3 & 58.3 & 71.4 & 67.3 \\
Qwen2.5-VL-7B & & \cmark & & & 56.4 & 85.3 & 64.3 & 60.7 & 71.3 & 67.6 \\
Qwen2.5-VL-7B & & & \cmark & & 56.2 & 85.2 & 63.5 & 60.6 & 71.0 & 67.3 \\
Qwen2.5-VL-7B & & & & \cmark & 56.8 & 85.6 & 64.3 & 60.4 & 71.1 & 67.6 \\
\rowcolor{lightgrayrow}Qwen2.5-VL-7B & \cmark & \cmark & \cmark & \cmark & 56.9 & 85.6 & 64.3 & 61.9 & 71.2 & 68.0 \\
\bottomrule
\end{tabular}}
\end{table*}
\begin{table*}[t]
\centering
\caption{Different evaluators for open-ended safety QA. LLM evaluators report larger gains than human evaluation, so the human result is interpreted as a conservative estimate.}
\label{tab:different_eval}
\tabfont
\setlength{\tabcolsep}{3pt}
\renewcommand{\arraystretch}{1.05}{%
\begin{tabular}{lc|cccc|cccc|cccc|cccc}
\toprule
\textbf{Model} & \textbf{Param.} & \multicolumn{4}{c|}{\textbf{GPT-4o}} & \multicolumn{4}{c|}{\textbf{Claude-3.7-Sonnet}} & \multicolumn{4}{c|}{\textbf{DeepSeek-R1}} & \multicolumn{4}{c}{\textbf{Human Eval}} \\
&& Acc. & Con. & Coh. & Overall & Acc. & Con. & Coh. & Overall & Acc. & Con. & Coh. & Overall & Acc. & Con. & Coh. & Overall \\
\midrule
Qwen2.5-VL~\cite{bai2025qwen2} & 3B & 57.2 & 70.2 & 70.3 & 61.1 & 59.3 & 73.5 & 75.6 & 63.8 & 50.7 & 76.8 & 77.5 & 58.6 & 71.3 & 69.4 & 71.9 & 70.9 \\
\rowcolor{lightgrayrow}Qwen2.5-VL + TSHA & 3B & 68.2 & 78.6 & 80.2 & 71.5 & 72.6 & 79.6 & 82.2 & 75.0 & 61.5 & 81.3 & 85.8 & 67.9 & 73.8 & 71.2 & 73.3 & 73.3 \\
\rowcolor{lightbluerow}$\Delta$ & 3B & +11.0 & +8.4 & +9.9 & +10.4 & +13.3 & +6.1 & +6.6 & +11.2 & +10.8 & +4.5 & +8.3 & +9.3 & +2.5 & +1.8 & +1.4 & +2.4 \\
\midrule
Qwen2.5-VL~\cite{bai2025qwen2} & 7B & 68.2 & 77.1 & 79.5 & 71.1 & 71.0 & 77.0 & 80.9 & 73.2 & 63.4 & 80.8 & 86.4 & 69.2 & 74.3 & 69.3 & 73.3 & 73.2 \\
\rowcolor{lightgrayrow}Qwen2.5-VL + TSHA & 7B & 71.6 & 80.2 & 81.8 & 74.3 & 76.2 & 80.7 & 83.8 & 77.9 & 65.6 & 82.6 & 87.6 & 71.2 & 76.3 & 71.8 & 75.1 & 75.3 \\
\rowcolor{lightbluerow}$\Delta$ & 7B & +3.4 & +3.1 & +2.3 & +3.2 & +4.8 & +2.3 & +2.9 & +4.7 & +2.2 & +1.8 & +1.2 & +2.0 & +2.0 & +2.5 & +1.8 & +2.1 \\
\midrule
Qwen2.5-VL~\cite{bai2025qwen2} & 32B & 73.6 & 79.5 & 82.5 & 75.7 & 75.9 & 76.5 & 82.3 & 76.7 & 68.1 & 79.2 & 86.1 & 72.1 & 81.9 & 74.6 & 78.7 & 80.1 \\
\rowcolor{lightgrayrow}Qwen2.5-VL + TSHA & 32B & 75.5 & 80.2 & 83.7 & 77.3 & 77.1 & 79.0 & 83.3 & 78.1 & 69.1 & 81.7 & 87.4 & 73.5 & 82.0 & 76.7 & 79.4 & 80.7 \\
\rowcolor{lightbluerow}$\Delta$ & 32B & +1.9 & +0.7 & +1.2 & +1.6 & +1.2 & +2.5 & +1.0 & +1.4 & +1.0 & +2.5 & +1.3 & +1.4 & +0.1 & +1.9 & +0.7 & +0.6 \\
\bottomrule
\end{tabular}}
\end{table*}

\subsection{Transfer to General Benchmarks}
\Cref{tab:other_bench} shows that TSHA yields positive average transfer for the tested Qwen2.5-VL sizes, with the largest average gain on 3B (+1.6) and 7B (+1.2). The gains are not uniform across all benchmarks. For 72B, the average gain is small (+0.4), with drops on BLINK (-2.2) and MMStar (-1.1) but gains on AI2D, MUIR, and SEED. We interpret this as a mild specialization trade-off: large models already possess strong general visual reasoning, and safety-focused fine-tuning can slightly shift decision boundaries away from some general benchmark distributions. This observation motivates reporting per-benchmark results rather than only average scores.

This transfer pattern is useful for understanding the scope of TSHA training. On smaller Qwen2.5-VL models, safety supervision appears to improve general visual grounding and instruction following, likely because the training requires the model to connect objects, spatial relations, and consequences. On the 72B model, the base model is already strong; additional safety-focused supervision produces modest average improvement but can slightly reduce performance on some general benchmarks. We therefore frame TSHA as a domain-specific safety resource that usually preserves general ability, not as a universal recipe for improving every multimodal benchmark.

The small BLINK and MMStar drops on 72B are informative because they represent the kind of trade-off practitioners need to know. Safety-hazard fine-tuning may make a model more sensitive to risk cues and more conservative in ambiguous scenes. This behavior is desirable for safety assessment but can shift answers on benchmarks where the expected reasoning distribution is different. For this reason, we recommend evaluating both the target safety domain and general VLM benchmarks when deploying TSHA-trained models.

\subsection{Source Ablation and Evaluator Robustness}
\Cref{tab:ablation} shows that each TSHA source contributes useful information. Internet images and newly captured images provide realistic domestic contexts; existing datasets provide broad indoor scene coverage; AIGC images supplement rare and severe hazards that are difficult or unsafe to stage, such as fire, smoke, electrical-water contact, and chemical risks. The full dataset performs best on average, supporting the multi-source construction design.

No single source dominates the ablation. Existing indoor datasets provide stable scene coverage and help models avoid overfitting to generated imagery. Internet frames contribute natural clutter, camera noise, and temporal-derived viewpoints. Newly captured images add volunteer-collected domestic layouts that are not necessarily common in public datasets. AIGC images expand the long tail of severe hazards. The best average transfer comes from combining all sources, which supports the design choice that TSHA should be heterogeneous rather than optimized around one data pipeline.

\Cref{tab:different_eval} directly addresses evaluator bias. GPT-4o reports a +10.4 gain for Qwen2.5-VL-3B, while human evaluation reports +2.4. This gap confirms that GPT-4o scores should not be treated as exact human-equivalent utility. Nevertheless, all three non-GPT evaluators and human raters agree on the direction of improvement for the tested Qwen models. We therefore claim robust directional improvement, not that GPT-4o provides an unbiased absolute score.

The evaluator comparison also explains why the paper reports several metrics rather than a single leaderboard number. LLM evaluators are useful for scaling open-ended evaluation, but they can differ in verbosity preference, hazard sensitivity, and tolerance for near-miss answers. Human evaluation is more conservative and better reflects practical correctness, but it is expensive to scale to every model and every sample. TSHA therefore uses automatic evaluation for broad coverage and human evaluation for calibration. The important result is not that all evaluators assign identical scores, but that the improvement direction remains stable across GPT-4o, Claude-3.7-Sonnet, DeepSeek-R1, and human evaluation.

\section{Conclusion}
\label{sec:conclusion}
We introduced TSHA, a benchmark for trustworthy indoor safety hazard assessment with broad scene coverage, a fine-grained hazard taxonomy, and challenging image, video, and panoramic test cases. TSHA exposes persistent weaknesses in current VLMs, especially for open-ended hazard identification. Training on TSHA substantially improves safety-hazard performance and yields modest average transfer to general VLM benchmarks, while also revealing model-size-dependent trade-offs. Beyond the dataset itself, TSHA contributes a transparent quality-control protocol covering annotation, human audit, AIGC plausibility, evaluator bias, and privacy filtering. We hope TSHA supports the development of VLMs that reason more reliably about everyday safety risks.

\bibliographystyle{IEEEtran}
\bibliography{main}

\end{document}